\documentclass{article}

\usepackage{microtype}
\usepackage{graphicx}
\usepackage{subfigure}
\usepackage{booktabs} 
\usepackage{bigints}
\usepackage{xcolor}
\usepackage{algorithm,algorithmic}

\usepackage{hyperref}
\usepackage{bm,amsmath}

\definecolor{gold}{rgb}{0.85,.66,0}
\definecolor{plum}{rgb}{0.45,0,.66}
\definecolor{mythistle}{rgb}{.99,.195,.133}
\definecolor{myred}{cmyk}{0.000000,1.000000,1.000000,0.1}
\definecolor{myblue}{cmyk}{1.000000,0.750000,0.000000,0.1}
\definecolor{mybgn}{cmyk}{0.850000,0.350000,0.000000,0.1}
\definecolor{mygrn}{cmyk}{0.750000,0.000000,1.000000,0.2}

\newcommand{\bl}{\begin{itemize}}
\newcommand{\el}{\end{itemize}}
\newcommand{\be}{\begin{enumerate}}
\newcommand{\ee}{\end{enumerate}}

\newcommand{\bea}{\begin{eqnarray*}}
\newcommand{\eea}{\end{eqnarray*}}

\newcommand{\beq}{\begin{equation}}
\newcommand{\eeq}{\end{equation}}

\newcommand{\bmx}{\left[ \begin{array}}
\newcommand{\emx}{\end{array} \right]}

\def\half{\frac{1}{2}}
\def\flip{\hat{\bf{x}}^c}

\newcommand{\mybf}[1]{\textbf{\em #1} }

\def\bfx{{\mybf{x}}}

\def\bfz{{\mybf{z}}}
\def\bfzed{{\bf{0}}}

\newcommand{\grad}{\bigtriangledown}

\DeclareMathOperator{\erf}{erf}
\DeclareMathOperator{\softmax}{softmax}
\def\bs#1{\boldsymbol{#1}}
\def\pij{\mbox{\scriptsize $(i,j)$}}

\newcommand{\rchange}{}
\newcommand{\dchange}{}
\newcommand{\echange}{}
\newcommand{\fchange}{}
\newcommand{\gchange}{}

\usepackage[accepted]{icml2019}

\icmltitlerunning{Interpreting Neural Networks Using Flip Points}

\begin{document}

\twocolumn[
\icmltitle{Interpreting Neural Networks Using Flip Points}



\icmlsetsymbol{equal}{}

\begin{icmlauthorlist}
\icmlauthor{Roozbeh Yousefzadeh}{umd}
\icmlauthor{Dianne P. O'Leary}{umda}
\end{icmlauthorlist}

\icmlaffiliation{umd}{Department of Computer Science, University of Maryland, College Park, USA}
\icmlaffiliation{umda}{\echange{Department of Computer Science and Institute for Advanced Computer Studies, University of Maryland, College Park, USA. oleary@cs.umd.edu}}

\icmlcorrespondingauthor{}{roozbeh@cs.umd.edu}

\icmlkeywords{Machine Learning, Explainable ML, Interpretability, Flip points}

\vskip 0.3in
]



\printAffiliationsAndNotice{}  

\begin{abstract}
Neural networks have been criticized for their lack of easy interpretation, which undermines confidence in their use for important applications. Here, we introduce a novel technique, interpreting a trained neural network by investigating its {\em flip points}. A flip point is any point that lies on the boundary between two output classes: e.g. for a neural network with a binary yes/no output, a flip point is any input that generates equal scores for ``yes" and ``no". The flip point closest to a given input is of particular importance, and this point is the solution to a well-posed optimization problem. \echange{This paper gives an overview of the uses of flip points and how they are computed.} Through results on standard datasets, we demonstrate how flip points can be used to provide detailed interpretation of the output produced by a neural network. Moreover, for a given input, flip points enable us to measure  confidence in the correctness of outputs much more effectively than $\softmax$ score. They also identify influential features of the inputs, identify bias, and find changes in the input that change the output of the model. We show that distance between an input and the closest flip point identifies the most influential points in the training data. Using principal component analysis (PCA) and rank-revealing QR factorization (RR-QR), the set of directions from each training input to its closest flip point provides explanations of how a trained neural network processes an entire dataset: what features are most important for classification into a given class, which features are most responsible for particular misclassifications, how an adversary might fool the network, etc. Although we investigate flip points for neural networks, their usefulness is actually model-agnostic.

\end{abstract}

\section{Introduction}

In real-world applications, neural networks are usually trained for a specific task and then used as a tool to perform that task, for example to make decisions or to make predictions. Despite their unprecedented success in performing machine learning tasks accurately and fast, these trained models are often described as black-boxes because they are so complex that one cannot interpret their output in terms of their inputs. 

When a trained network is used as a black-box, users cannot be sure how confident they can be in the correctness of each individual output. Furthermore, when an output is produced, it would be desirable to know the answer to questions such as, what changes in the input could have made the output different? A black-box cannot provide answers to such questions. This inexplainability becomes problematic in many ways, especially when the network is utilized in tasks consequential to human lives, such as in criminal justice, medicine, and business. Because of these interpretation issues, there have been calls for avoiding neural networks in high-stakes decision making \cite{rudin2018please}. Alternatives include Markov decision processes \cite{lakkaraju2017learning}, scoring systems \cite{rudin2018optimized, chen2018interpretable}, binary decision trees \cite{Bertsimas2017}, and Bayesian rule sets \cite{wang2016bayesian}.

There have been several approaches for interpreting neural networks and general black-box models. We have space here to mention only a few papers representative of the field.

Some studies have taken a model-agnostic approach to interpreting black box models such as neural networks. For example, the approach taken by \citet{ribeiro2016should}  builds an explanation for an output via a linear model in the vicinity of a specific input. Similarly, \citet{anchors:aaai18} derive if-then rule explanations about the local behavior of black box models.

Methods based on perturbing each input feature individually have severe computational limitations. First, they can be prohibitively expensive when dealing with a complex high-dimensional nonlinear function such as that represented by a neural network. Second, the output of a neural network can be constant over vast areas of its domain, while it might be very volatile in other regions. Therefore, it can be hard to find a suitable vicinity that gives sensible results when perturbing high-dimensional inputs. Third, the features may have incompatible scalings, so determining meaningful perturbations is difficult. Finally, the features of the inputs can be highly correlated; therefore, perturbing the inputs one by one will be inefficient and possibly misleading. \citet{koh2017understanding} have used influence functions to guide the perturbation and interpret black-box models with emphasis on finding the importance of individual points in the training data.


Pursuing the interpretation of neural networks from an adversarial point of view, \citet{ghorbani2017interpretation}  generate adversarial perturbations that produce perceptively indistinguishable inputs that are assigned the same label, yet have very different interpretations. They further show that interpretations based on exemplars (e.g. influence functions) are similarly susceptible to adversarial attack.

Another line of research focuses on performing insightful pre-processing to make the inputs to the neural network more interpretable. One promising approach uses prototypes to represent each output class \rchange{\cite{li2017deep,chen2018looks,snell2017prototypical}}. Individual inputs are compared to the prototypes (e.g., by measuring the 2-norm distance between each input and all the prototypes), and that information is the input to the neural network. In the context of text analysis, \citet{lei2016rationalizing} has introduced a model that first specifies distributions over text fragments as candidate rationales and then uses the rationales to make predictions. 



Taking a different approach, \citet{lakkaraju2017interpretable} have used decision rules to emulate a neural network in a subdomain of the inputs. Although the emulated model in their numerical example is interpretable, its outputs are different than the outputs of neural network for about 15\% of the data.

\rchange{Most recently, \cite{spangher2018actionable} have proposed the idea of using \textit{flip sets} to study linear classifiers \dchange{(logistic regression models, linear support vector machines, etc.)}. They define a flip set as the \dchange{set} of changes in the input that can flip the prediction of a classifier. This approach has similarities to ours, however their \dchange{algorithm} is only applicable to linear classifiers} \dchange{and does not handle the nonlinearities in neural networks.}



Many alternative models such as decision trees and rule lists have been in competition and co-existence with neural networks for decades, but in many applications have not been very appealing with respect to accuracy, scalability, and complexity, particularly with high-dimensional data. 
Our goal is to improve the interpretability of neural networks and other black-box models so that in cases where they have computational or accuracy advantages over alternative models, they can be used without hesitation.
Through the use of {\em flip points} we are able to 
make neural networks interpretable, improve their training, and indicate the reliability of the output classification.

\dchange{\comment{\section{Making neural networks explainable} 
\label{sec-make-nn-exp}

-\rchange{I think it might be good to drop Section 2 entirely and use it in the homotopy paper. I think we raise detailed concerns in the reader's mind, but we do not have space to adequately address those concerns.}

A trained neural network enables users such as medical professionals and judges to make decisions, make predictions, and solve problems. Interpretation relates output to input in a way meaningful to the user and answers questions such as what data led to a particular recommendation.

Viewing the neural network as a mathematical function provides highly useful tools for interpretation. For example, for a neural network that makes binary predictions on whether patients have cancer or not, we demonstrate below that we can formulate an optimization problem that finds the least change in the features of a tumor predicted to be ``non-cancerous" that would cause the model to change the prediction to ``cancerous''. 

Unfortunately, there are two factors making this optimization problem quite difficult to solve.
First, each output of a neural network is a \underline{nonconvex} function of the inputs, formed by hierarchical composition of many, many functions. For this reason, we can generally find a close point but not necessarily the closest.
Second, the output is often \underline{nearly constant} over vast regions of its domain. In these regions, the derivatives will be close to zero, meaning that methods based on gradient descent move in an essentially random direction, determined by round-off in the computed derivatives. In practice, we overcome this problem using a homotopy method, embedding the original neural network  in a family of related networks and using them to guide us to a solution, as we explain in forthcoming work.
}}

\vspace{-2mm}

\section{Interpreting neural networks using flip points} \label{sec_define_flip}

In this section, we first study the interpretation of a neural network that has two outputs and then extend the results to neural networks with an arbitrary number of outputs.

We consider a general neural network with multiple layers. 
We assume that the activation function at each node is a continuous function of its inputs, a condition satisfied by all commonly used activation functions. 
For notation, we use $\bfx$ for the vector of inputs to the neural network and $\bfz$ for the vector of outputs.

\dchange{\echange{Here we} define the optimization problems used to compute flip points. Algorithms for the numerical solution of the problems are discussed in Appendix A.}

\subsection{Neural networks with two outputs: a binary prediction}

First, consider a neural network with two output nodes. For definiteness, we refer to the output of the neural network as a prediction of ``cancerous" or ``noncancerous", but our results are equally applicable to other types of output, such as decisions. We assume that the output $\bfz(\bfx)$ is normalized (perhaps using $\softmax$) so that the two elements of the output sum to one. Since $z_1(\bfx) + z_2(\bfx) = 1$, we can specify the prediction by a single output: $z_1(\bfx) > \half$ is a prediction of ``cancerous", and $z_1(\bfx) < \half$ is a prediction of ``noncancerous". If $z_1(\bfx) = \half$, then the prediction is undefined.

Now, given a prediction $z_1(\bfx) \ne \half$ for a particular input $\bfx$, we want to investigate how changes in $\bfx$ can change the prediction, for example, from ``cancerous" to ``noncancerous". In particular, it would be very useful to find the {\it least change} in $\bfx$ that makes the prediction change. 

Since the output of the neural network is continuous, $\bfx$ lies in a region of points whose output $z_1$ is greater than $\half$, and the boundary of this region is continuous. So what we really seek is a nearby point on that boundary, and we call points on the boundary {\it flip points}.
So given $\bfx$ with $z_1(\bfx) > \half$, we seek a nearby point $\hat{\bfx}$ with $z_1(\hat{\bfx}) = \half$.\footnote{
One technical point: Because $z_1$ is continuous, there will be a point arbitrarily close to $\hat{\bfx}$ for which $z_1$ is less than $\half$ and the prediction becomes ``noncancerous" {\it unless} $\hat{\bfx}$ is a local minimizer of the function $z_1$. In this extremely unlikely event, we will have the gradient $\grad z_1(\hat{\bfx}) = \bfzed$ and the second derivative matrix positive semidefinite, and $\hat{\bfx}$ will not be a boundary point. In practice, this is not likely to occur.}

The closest flip point $\flip$ is the solution to an optimization problem
\begin{equation} \label{eq-2o-obj}
    \min\limits_{\hat{\bfx} } \| {\hat{\bfx}} - \boldsymbol{x} \| ,
\end{equation}
where $\| . \|$ is a norm appropriate to the data.
Our only constraint is
\[
z_1(\hat{\bfx} )= \half.
\]
Specific problems might require additional constraints; e.g., if $\boldsymbol{x}$ is an image, upper and lower bounds might be imposed on $\hat{\bfx}$, and discrete inputs will require binary or integer constraints.
It is possible that the solution $\flip$ is not unique, but the minimal distance is always unique.

\subsection{Interpreting neural networks with multi-class outputs}

For neural networks with multi-class outputs, we can use this same approach to define flip points between any pair of classes and to find the closest flip points for a given input.  Suppose our neural network has $n_z$ outputs and, for $\bfx$, $z_i(\bfx)$ is the largest component of $\bfz(\bfx)$. If we want to find a flip point between classes $i$ and $j$, then
the objective function \eqref{eq-2o-obj} remains the same, and the constraints become
\[
z_i(\hat{\bfx}) = z_j(\hat{\bfx}),
\]
and, for $k \ne i,j$,
\[
z_i(\hat{\bfx}) > z_k(\hat{\bfx}).
\]
Thus, for each individual input, we can compute $n_z - 1$ closest flip points \rchange{$\flip\pij$} between the class for that input and each of the other classes.

\subsection{Interpreting neural networks with a quantified output}
 Neural networks can also be used to specify a quantity. For example, a neural network can be trained to determine the appropriate dosage of a medicine. In such applications, flip points have a different meaning. For example, we can ask for the least change in the input that changes the dose by a given amount. Again, we can formulate and answer these questions as optimization problems. This will be the subject of future work.

\section{How flip points provide valuable information to the user}

We now explain how flip points can be used to improve the performance and interpretation of neural networks.

\subsection{Determine the least change in $\bfx$ that alters the prediction of the model} \label{sec:leastch}
The vector $\flip - \bfx$ is an accurate and clear explanation of the minimum change in the input that can make the outcome different. This is insightful information that can be provided along with the output. For example, in a bond court, a judge could be told what changes in the features of a particular arrestee could produce a ``detain" recommendation instead of a ``release" recommendation.

\subsection{Assess the trustworthiness of the classification for $\bfx$}

In our numerical examples we show that the numerical value of the output of a neural network, when the last layer is defined by the $\softmax$ function, does not indicate how sure we should be of the correctness of the output. In fact, many mis-predictions correspond to very high $\softmax$ values.
\dchange{This has been previously observed \cite{nguyen2015deep,guo2017calibration}. 
\citet{gal2016dropout} propose using information from training using dropout to assess the uncertainty of predictions. Their method is restricted to this particular training method, does not provide the likely correct prediction, and is more expensive than the method we propose.
Another approach, proposed by \citet{guo2017calibration} constructs a calibration model, trained separately on a validation set, and appends it as a post-processing component to the network. 
Also, \citet{lakshminarayanan2017simple}  used ensembles of neural networks, trained adversarially with pre-calculated scoring rules, in order to estimate the uncertainty in predictions.
Using flip points to assess the trustworthiness of predictions is a novel idea that has certain advantages compared to other approaches in the literature, as we explain.
}

\dchange{The} distances of incorrectly classified points to their flip points tend to be very small compared to the distances for correct predictions, implying that closeness to a flip point is indicative of how sure we can be of the correctness of a prediction. Small distance to the closest flip point means that small perturbations in the input can change the prediction of the model, while large distance to the flip point means that a larger change is necessary. It is important, of course, that distance be measured in a meaningful way, with input features normalized and weighted in a way that emphasizes their importance. \dchange{Furthermore, in multi-class predictions, our numerical results indicate that when the model makes an incorrect classification, the class with the closest flip point is actually the correct class.}


\rchange{Using flip points can be viewed as a direct method to assess the trustworthiness of predictions, even when models are calibrated or trained adversarially. Therefore, flip point assessment is not necessarily in competition with \dchange{other} methods in the literature; rather it is a simple and straightforward method that can be used for  \dchange{any} model. Flip points also provide clear \dchange{explanations} for their assessment in terms of input features and can point out to the possible correct prediction when there is low confidence.
}

\subsection{Identify uncertainty in the classification of $\bfx$}

Often, some of the inputs to a neural network are measured quantities which have associated uncertainties. When the difference between $\bfx$ and its closest flip point is less than the uncertainty in the measurements, then the prediction made by the model is quite possibly incorrect, and this information should be communicated to the user.

\subsection{Use PCA analysis of the flip points to gain insight about the dataset}
In Section \ref{sec:leastch} we discussed using the direction from a single data point to the closest flip point to provide sensitivity information.
Using PCA analysis, we can extend this insight to an entire dataset or to subsets within a dataset,

We form a matrix with one row $\flip - \bfx$  for each data point. PCA analysis of this matrix identifies the most influential directions for flipping the outputs in the dataset and thus the most influential features,
This procedure provides clear and accurate interpretations  of the neural network model. One can use nonlinear PCA or auto-encoders to enhance this approach.

Alternatively, for a given data point, PCA analysis of the directions from the data point to a collection of boundary points can give insight about the shape of the decision boundary.

\section{How flip points can improve the training and security of the neural network}

Flip points also provide valuable information that can improve the quality and efficiency of the training process.

\subsection{Identify the most and least influential points in the training data in order to reduce training time}

Points that are correctly classified and far from their flip points have little influence on setting the decision boundaries for a neural network. Points that are close to their flip points are much more influential in defining the boundaries between the output classes. 
    
Therefore, in online learning and real-time applications, where we have to retrain a neural network using streaming data, we can retrain the network more quickly using only the influential data points, those with small distance from their flip points.

\rchange{\dchange{As mentioned earlier, \citet{koh2017understanding}  use influence functions to relate individual predictions of a trained model to training points that are most influential for that prediction.}
\dchange{They are} not able to draw conclusions about the decision boundaries of the model \dchange{because they use} small perturbations of training data and \echange{local gradient information for} the loss function, which can be misleading \dchange{for} nonlinear non-convex functions in high dimensional space. Our approach \dchange{does not just rely on local information but it seeks} the closest point that flips the decision of \dchange{the} network. Therefore, the insight we provide goes well beyond their method  \dchange{without adding prohibitive expense}.

}

\rchange{
\subsection{\dchange{Identify} out-of-distribution points in the data and \dchange{investigate} overfitting}
\dchange{Out-of-distribution points in the training set appear as incorrectly classified points with large distance to the closest flip point.}
\dchange{Finding} such points \dchange{can identify errors in the input or} subgroups in the data that do not have adequate representation in the training set (e.g., faces of people from a certain race in a facial recognition dataset \cite{buolamwini2018gender}). 

Additionally, after we compute the closest flip points for all the points in the training set, we can further cluster the flip points and study each cluster in relation to its nearby data points. This will potentially enable us to investigate whether the model has overfitted to the data points or not. 

We have not investigated these two opportunities in our numerical results, \dchange{but} believe \dchange{that they are} promising directions \dchange{for study}.
}

\subsection{Generate synthetic data to improve accuracy\rchange{ and to shape the decision boundaries}}

We can use flip points as {\it synthetic data}, adding them to the training set to move the output boundaries of a neural network insightfully and effectively.

Suppose that our trained neural network correctly classifies a training point $\bfx$ but that there is a nearby flip point $\flip$. We generate a synthetic data point by adding $\flip$ to the training set, using the same classification as that for $\bfx$. Retraining the network will then tend to push the classification boundary further away from $\bfx$.

Similarly, if our trained neural network makes a mistake on a given training point $\bfx$, then we can add the flip point $\flip$ to the training set, giving it the  same classification as $\bfx$. This reinforces the importance of the mistake and tends to correct it.

\rchange{Using flip points to alter the decision boundaries can be performed not just to improve the accuracy of a model but also to change certain traits adopted by the trained network. For example, if a model is biased \dchange{for} or against certain features of the inputs, we could alter that bias using synthetic data. We will demonstrate this later in our numerical results on the Adult Income dataset.

There are studies in the literature that have used synthetic data (other than flip points) to improve the accuracy, e.g., \cite{jaderberg2014synthetic}. There is also a line of research that \echange{has} used perturbations of the inputs in order to make the trained models robust, \citet{tsipras2018robustness} for example.
However, the idea of using the \textit{flip points} as \textit{synthetic data} is novel and would benefit the studies on robustness of networks, too.
}

\subsection{Understand adversarial influence}
Flip points also provide insight for anyone with adversarial intentions. First, these points can be used to understand and exploit possible flaws in a trained model. Second, adding flip points with incorrect labels to the training data will distort the class boundaries in the trained model and can diminish its accuracy or bias its results. Our methods could be helpful in studying adversarial attacks such as the problems studied by \citet{schmidt2018adversarially}, \citet{sinha2018certifying}, \citet{madry2017towards}, \rchange{and \citet{katz2017towards}}.

\section{Results} \label{sec_results}

In our numerical results, we use feed-forward neural networks with 12 layers and $\softmax$ on the output layer. We use a tunable error function as the activation function and use Tensorflow for training the networks, with Adam optimizer and learning rate of 0.001. \rchange{Keep in mind that one can compute the flip points for trained models and interpret them, regardless of the architecture of the model (number of layers, activation function, etc.), the training set, and the training regime (regularization, etc.).}

When calculating  flip points, we measure the distance in equation \eqref{eq-2o-obj} using the 2-norm. \rchange{Calculating the closest flip points is quite fast, under 1 second for the MNIST, CIFAR-10, and Wisconsin Breast Cancer datasets using a 2017 MacBook. Calculating the closest flip point for the Adult Income dataset takes about 5 seconds, because it has both discrete and continuous variables.}

\subsection{Image classification}
\subsubsection{MNIST}
The MNIST dataset has 10 output classes, corresponding to the digits 0 through 9. We 
\dchange{could use pixel data as input to the networks, but, for efficiency, we choose to} represent each data point as a vector of length 100, using the Haar wavelet basis. The 100 most significant wavelets are chosen by rank-revealing QR decomposition \cite{chan1987rank} of the matrix formed from the wavelet coefficients of all images in the training set. 
\dchange{The wavelet transformation is a systematic way of applying convolutions of various widths to the input data, and the reduction applied by using rank-revealing QR decomposition leads to significant compression of the input data, from \rchange{784} features to 100, allowing us to use smaller networks. This idea, independent of flip points, is valuable whenever working with image data.}
\rchange{Using pixel input instead of wavelet coefficients would yield interpretation traits similar to those that we present here.}

We train two networks, NET1 and NET2, using half of the training data (30,000 images) for each. 
Table~\ref{table-mnist1} shows the accuracy of each network in the 2-fold cross validation.
Accuracy could be improved using techniques such as skip architecture, but these networks are adequate for our purposes.


\begin{table}[th]
\caption{Classification accuracies for NET1 and NET2 for MNIST.}
\label{table-mnist1}
\vskip 0.15in
\begin{center}
\begin{small}
\begin{sc}
\begin{tabular}{  p{1.4cm}  p{1.75cm}  p{1.8cm}  p{1.6cm} }
\toprule
Trained network & Accuracy on 1st half of training set & Accuracy on 2nd half of training set & Accuracy on testing set \\
\midrule
net1    & 100\% & 97.62\% & 97.98\% \\
net2    & 97.56\% & 100\% & 97.64\% \\
\bottomrule
\end{tabular}
\end{sc}
\end{small}
\end{center}
\vskip -0.1in
\end{table}

For each of the images in the training set, we calculate the flip points between the class predicted by the trained neural networks and each of the other 9 classes. %

\begin{figure}[ht]
\vskip -0.1in
\centering
\centerline{\includegraphics[width=0.6\columnwidth]{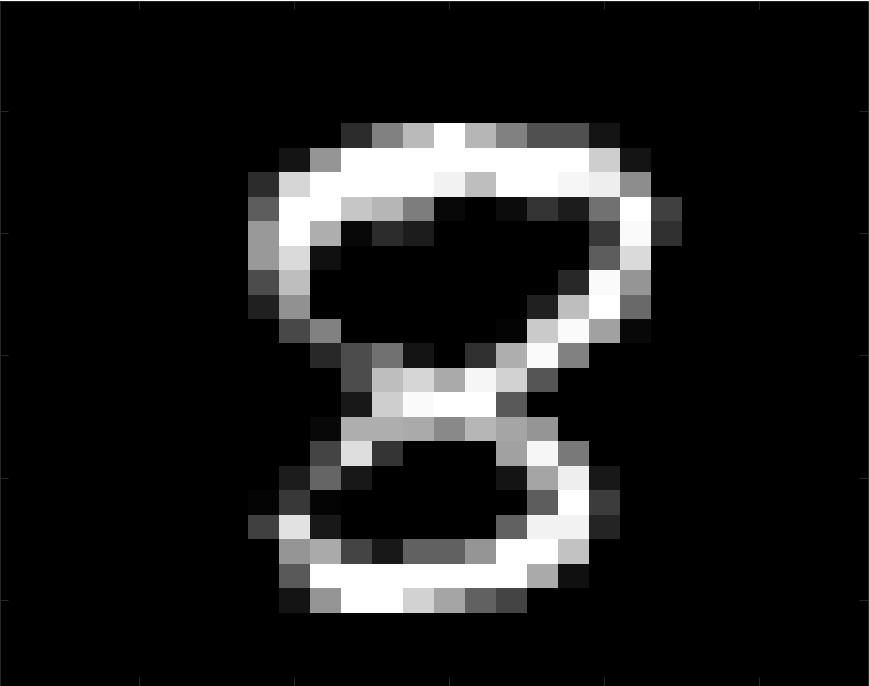}}
\caption{MNIST image mistakenly classified as ``3" by NET1.}
\label{fig-mnist8}
\vskip -0.1in
\end{figure}

{\bf Flip points identify alternate classifications.} Some images are misclassified and close to at least one flip point. For all of these points, the correct label is identified by the closest of the 9 flip points (or one of those tied for closest after rounding to 4 decimal digits).
For example, the image shown in Figure \ref{fig-mnist8}, from the second half of the MNIST training set,  is an ``8"  mistakenly classified as ``3" by NET1 with $\softmax$ score of 98\%. Its distances to the closest flip points  are shown in Table \ref{table-mnist2}. Assuming that we do not know the correct label for this image, we would report the label as ``3", with the additional explanation that there is low confidence in this prediction (because of closeness to the flip point), and the correct label might be ``8".




\begin{table}[h]
\setlength{\tabcolsep}{3.5pt}
\caption{Distance to closest flip points between class ``3" and other classes, for image in Figure \ref{fig-mnist8}.}
\label{table-mnist2}
\vskip 0.15in
\begin{center}
\begin{footnotesize}
\begin{sc}
\begin{tabular}{  cccccccccc }
\toprule
Class & 0 & 1 & 2 & 4 & 5 & 6 & 7 & 8 & 9\\
\midrule
Distance & \scriptsize{1.27}  & \scriptsize{1.32} & \scriptsize{0.58} & \scriptsize{2.16} & \scriptsize{0.56} & \scriptsize{1.45} & \scriptsize{1.51} & \scriptsize{0.16} & \scriptsize{0.90} \\
\bottomrule
\end{tabular}
\end{sc}
\end{footnotesize}
\end{center}
\vskip -0.1in
\end{table}

\begin{figure}[ht]
\vskip -0.1in
\centering
\centerline{\includegraphics[width=1.15\columnwidth]{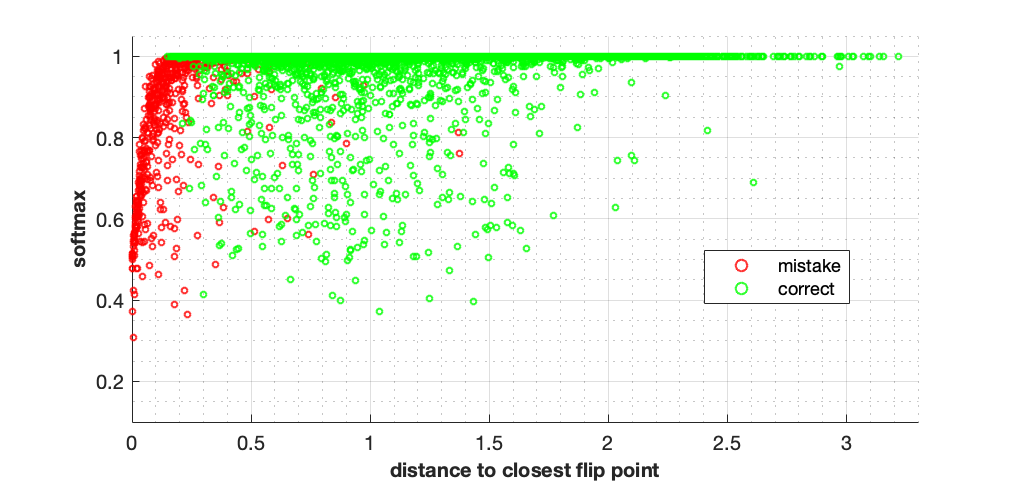}}
\caption{For the MNIST data, a large $\softmax$ score says nothing about the reliability of the classification. In contrast,  distance to the closest flip point is a much more reliable indicator.}
\label{fig-mnist-soft}
\vskip -0.1in
\end{figure}

\begin{figure}[ht]
\centering
\centerline{\includegraphics[width=0.98\columnwidth]{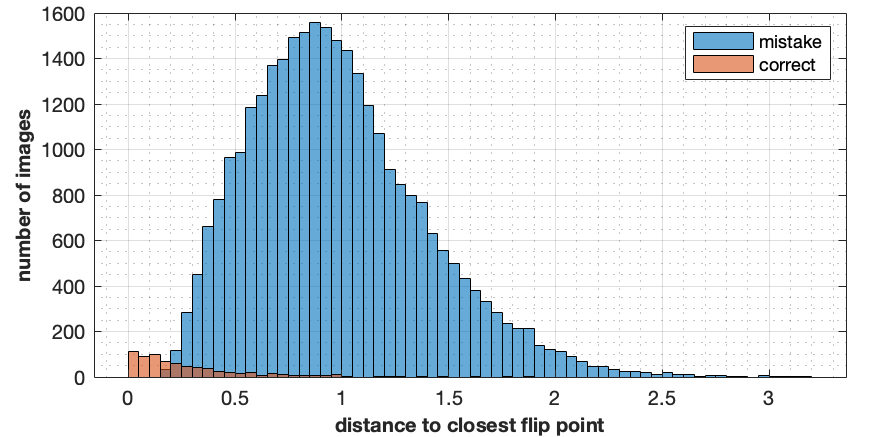}}
\caption{Distribution of distance to closest flip point among the images in the MNIST training set for mistakes (orange) and correctly classified points (blue).}
\label{fig-mnist-dist}
\vskip -0.1in
\end{figure}

{\bf Flip points provide better measure of confidence than softmax.} Many practitioners use the $\softmax$ output as a measure of confidence in the correctness of the output. As illustrated in Figure~\ref{fig-mnist-soft} , the $\softmax$ scores range between 31\% and 100\% for the mistakes by NET1 and NET2, and range between 37\% and 100\% for correct classifications, providing no separation between the groups. If $\softmax$ were a good proxy for distance, then the data would lie close to a straight line. Instead, most of the mistakes have small distance but large $\softmax$ score: more than 73\% of the mistakes have 80\% or more $\softmax$ score. Hence, $\softmax$ cannot identify mistakes. Fortunately, the figure shows that the distance to the closest flip point is  a much more reliable indicator of mistakes: mistakes almost always correspond to small distances. This is further demonstrated in Figure~\ref{fig-mnist-dist} which shows the distinct difference between the distribution of distances for the mistakes and the distribution of distances for  the correct classifications.


{\bf Flip points identify influential training points.}  Images that are correctly classified but are relatively close to a flip point are the most influential ones in the training process. To verify this, consider  the first half of the MNIST training set, and order the images by their distances to their nearest $\flip$ for NET1. We then consider using neural networks trained using a subset of this data.

Data points at most  $0.75$ from a flip point form a subset of $9,463$ images, about 15\% of the training set. A model trained on this subset achieves 97.9\% accuracy on the testing set. When we train with a subset of $9,463$ images randomly chosen from the training set, on average (50 trials) we achieve 96.2\% accuracy on the testing set. A subset of same size from the images farthest from their flip points achieves only  90.6\%  accuracy on the testing set. 

\begin{figure}[ht]
\vskip -0.1in
\centering
\centerline{\includegraphics[width=1.15\columnwidth]{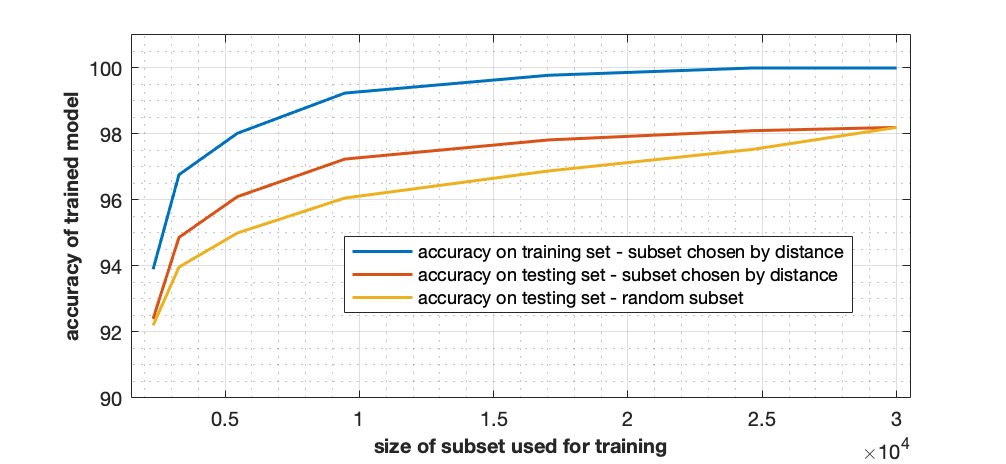}}
\caption{Accuracy of models trained on MNIST subsets.}
\label{fig-mnist-subset}
\vskip -0.1in
\end{figure}

These trends hold for all distance thresholds, as shown in Figure~\ref{fig-mnist-subset}. This confirms that distance to the flip point is in fact related to influence in the training process.

We note that the model learns the entire training set with 100\% accuracy when trained on about 16,000 images chosen by the distance measure. In contrast, it only achieves 98.8\% accuracy when trained on a randomly chosen subset of the same size.

Also note that flip points are computed by solving a non-convex optimization problem, so we cannot guarantee that we have indeed found the {\em closest} flip point. Nevertheless, in practice, the computation seems to provide very {\em useful} flip points, validated by the small distances achieved by some flip points and by the results shown in Figures~\ref{fig-mnist-soft}~--~\ref{fig-mnist-subset}.

{\bf Flip points improve the training of the network.}
Finally, we append to the entire training set a flip point for each mistake in the training set, labeled with the correct label for the mistake. The resulting neural network  achieves 100\% accuracy on the appended training set and 98.6\% on the testing set, an improvement over the 98.2\% accuracy of the original network.   We expect that this technique of appending  synthetic images to the training set will be much more helpful for datasets that have a limited amount of training data.



\subsubsection{CIFAR-10}
We now consider two classes of airplanes and ships in the CIFAR-10 data set. This time we perform 3D wavelet decomposition on images using the Haar wavelet basis and use all of the wavelet coefficients to train a neural network, achieving 100\% and 84.2\% accuracy on the training and testing sets. We then calculate the flip points for all the images in both sets.

\dchange{Observations that we  reported for MNIST on CIFAR-10 apply here, too. So, we focus our discussion on the directions to flip points and PCA analysis of them.}

\begin{figure}[ht]
\vskip -0.1in
\centering
  \begin{minipage}[b]{0.49\columnwidth}
\centerline{\includegraphics[width=1\columnwidth]{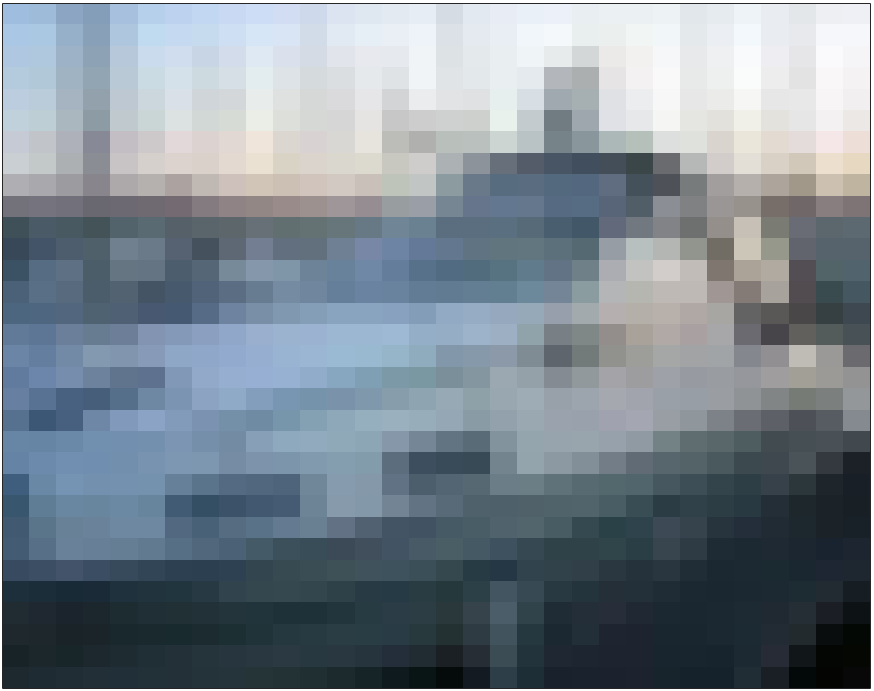}}
  \end{minipage}
  \begin{minipage}[b]{0.49\columnwidth}
\centerline{\includegraphics[width=1\columnwidth]{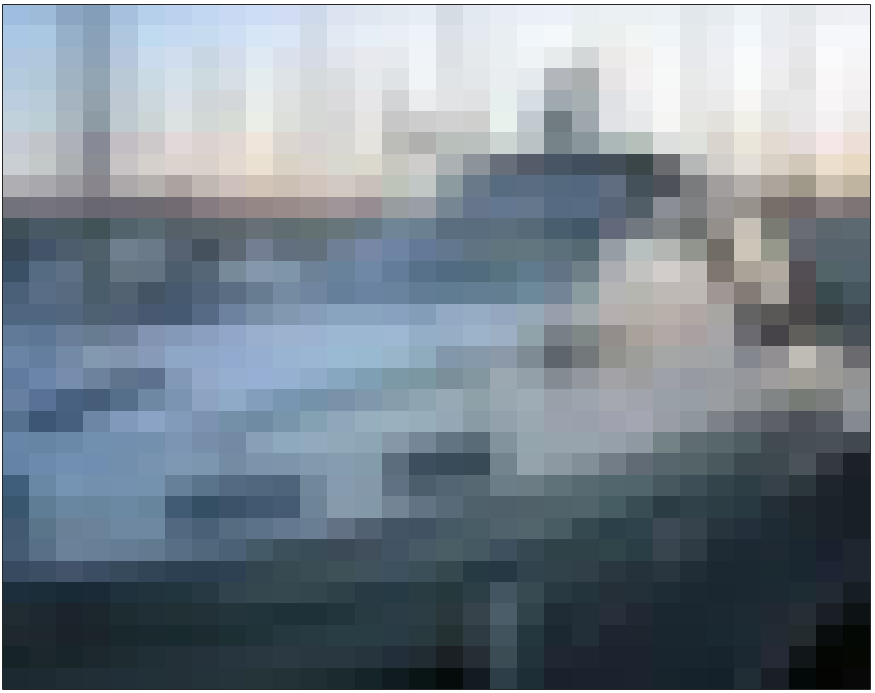}}
  \end{minipage}
\caption{A ship image misclassified as airplane (left), its flip point (right).}
\label{fig_cifar_ship}
\vskip -0.1in
\end{figure}

Figure \ref{fig_cifar_ship} shows an image in the testing set that is mistakenly classified as an airplane, along with its closest flip point. \rchange{We have computed the closest flip point in the wavelet space. It is interesting that t}he 1-norm distance between the image and its closest flip point \rchange{in the pixel space} is 210, and the differences are hard to detect by eye.

The matrix of directions between the misclassified images and their closest flip points is highly rank deficient. While we have 2,304 features for each image, the rank of directions for flipping an airplane to a ship is 162, and it is 170 for flipping a ship to an airplane. Therefore, we can investigate the mistakes by looking at very small \rchange{subset} of wavelet features \rchange{out of the 2304 features}.

\begin{figure}[ht]
\vskip -0.1in
\centering
\centerline{\includegraphics[width=.6\columnwidth]{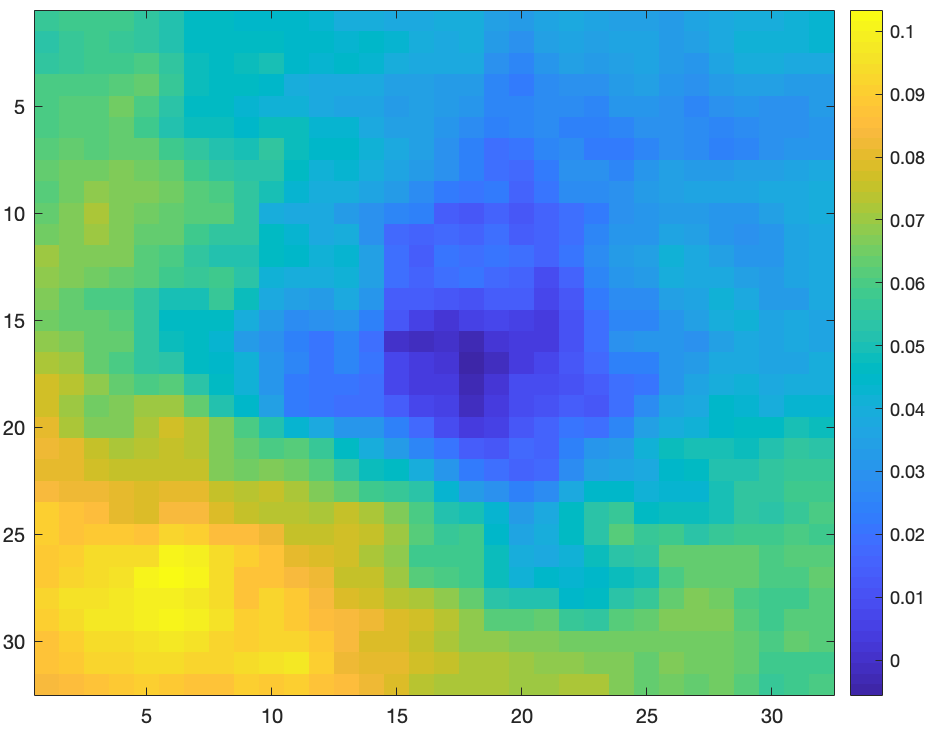}}
\caption{First principal component of directions that flip a misclassified ship to its correct class.}
\label{fig_cifar_pca_ship}
\vskip -0.1in
\end{figure}

Moreover, the matrix of directions that flip a misclassified ship to its correct class has 53\% sparsity. The first principal component of the directions has the pattern shown in Figure~\ref{fig_cifar_pca_ship}. 

\begin{figure}[ht]
\vskip -0.1in
\centering
  \begin{minipage}[b]{0.24\columnwidth}
\centerline{\includegraphics[width=1\columnwidth]{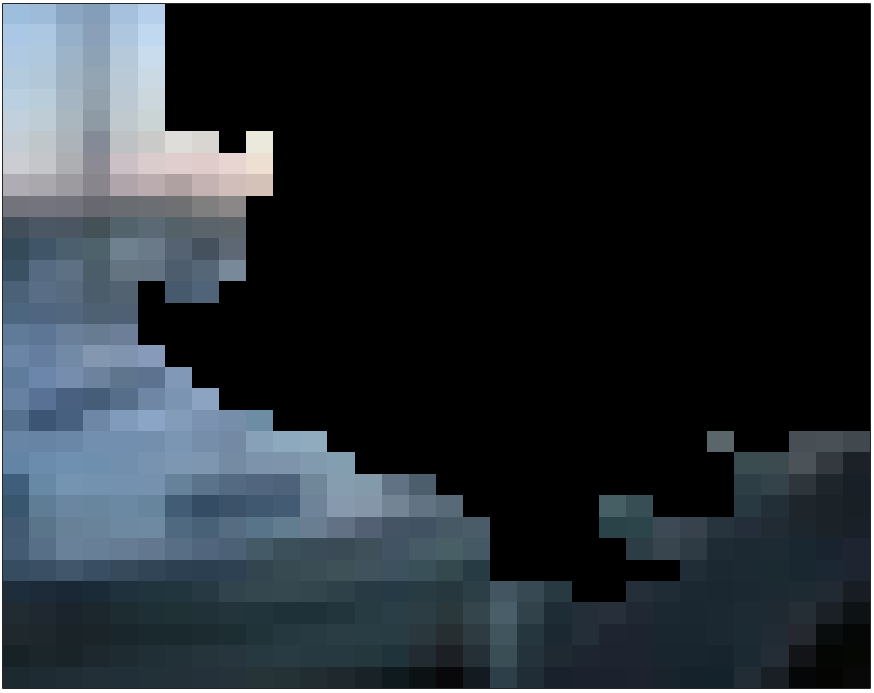}}
  \end{minipage}
  \begin{minipage}[b]{0.24\columnwidth}
\centerline{\includegraphics[width=1\columnwidth]{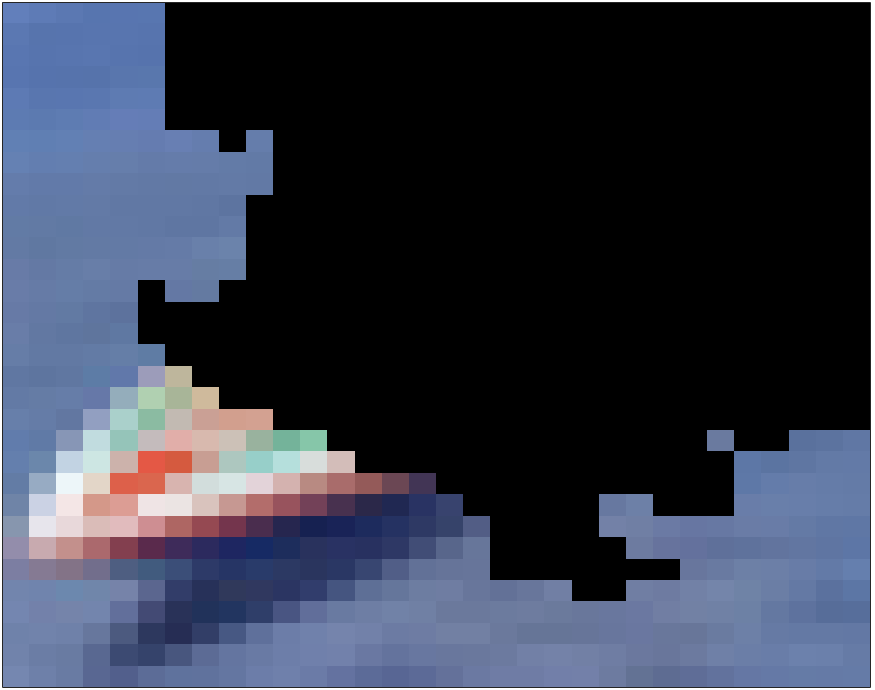}}
  \end{minipage}
  \begin{minipage}[b]{0.24\columnwidth}
\centerline{\includegraphics[width=1\columnwidth]{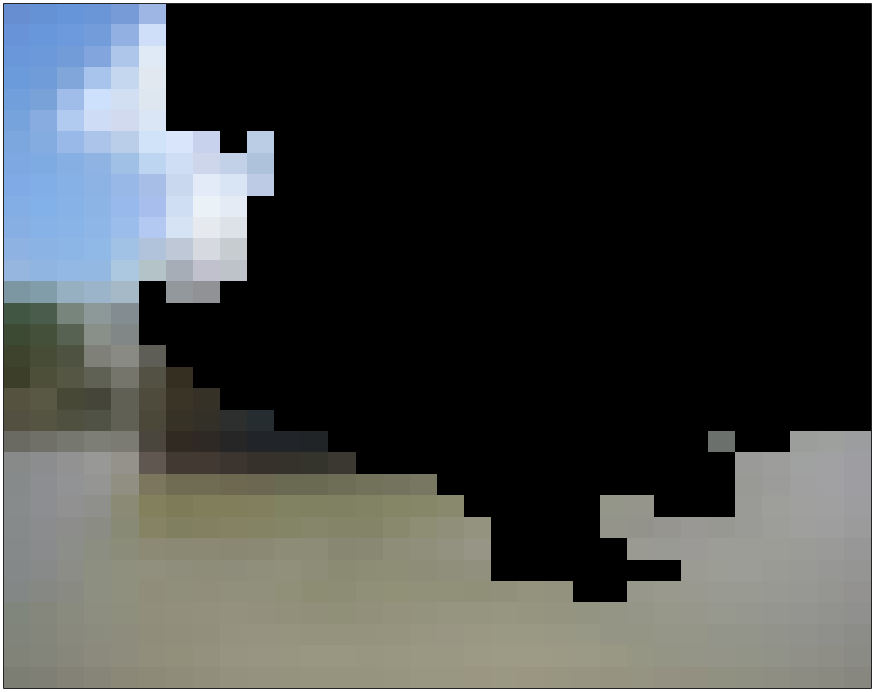}}
  \end{minipage}  
  \begin{minipage}[b]{0.24\columnwidth}
\centerline{\includegraphics[width=1\columnwidth]{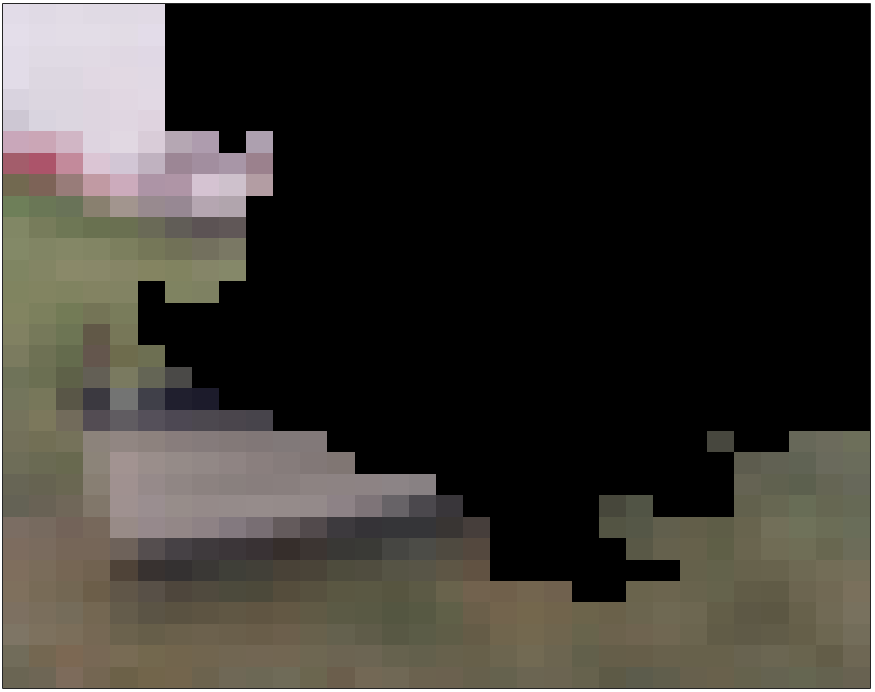}}
  \end{minipage}
\caption{Pixels with large principal coefficients for misclassified ships.}
\label{fig_cifar_pixels}
\vskip -0.1in
\end{figure}

We threshold the principal coefficients in Figure~\ref{fig_cifar_pca_ship}, retaining pixels with coefficient greater than 0.05. Then we  plot the corresponding pixels of the misclassified images of ships. Some of those images are plotted in Figure~\ref{fig_cifar_pixels}. One can see that for many of the mistakes, those pixels actually contain the prow of the ship in the image. This points to one vulnerability of our trained neural network,  which we could then investigate further.

Finally, we note that great similarity exists between the directions for the correct classifications in the training and testing sets. \echange{Investigating other principal components can provide additional insights.}



\subsection{Adult income dataset}
Next, we consider the Adult dataset from the UCI Machine Learning Repository \cite{Dua2017}, an example that has a combination of discrete and continuous variables.  There are 32,561 data points in the training set and 16,281 in the testing set. Each data point has information about an individual, and the label is binary, indicating whether the individual's income is greater than 50K annually.

Each of the continuous variables (age, fnlwgt, education-num, capital-gain, capital-loss and hours-per-week) has a lower bound of 0.  We normalize each variable to the range 0 -- 100 using upper bounds of 100, 2e6, 20, 2e5, 1e4 and 120, respectively, and we also use these ranges to constrain the search for flip points. 

Moreover, we transform the categorical variables (workclass, education level, marital status, occupation, relationship, race, sex, native country) into a binary form where each category type is represented by one binary feature. The categories that are active for a data point have binary value of 1 in their corresponding features, while the rest of features are set to zero. 
When searching for a flip point, we have a constraint that requires exactly one binary element be equal to 1 for each of the categorical variables. Our trained neural network achieves accuracy of 87.3\% and 86.1\%  on the training and testing sets.

Our aim here is to show how a trained neural network can be interpreted, and our focus is not to draw conclusions about the dataset itself. Clearly, choosing a different method for pre-processing the data, choosing different bounds for continuous variables, or defining other distance measures between the categories can affect the trained neural network function and consequently the result of interpretation. Our interpretation tools allow the user of model to become insightful about how to process the data and train the network in order to obtain a trained neural network whose output obeys known relationships between input and output. \dchange{Our preprocessing and scaling choices are suboptimal but illustrative; clearly, application scientists should always be involved in setting the distance metric in order to ensure meaningful results.}

{\bf Flip points provide interpretations and can expose bias.} 
To illustrate the interpretation of the output of the neural network, consider the 53rd training data point, corresponding to a person with income greater than 50K. The answer to the question of how features differ between this point and its computed flip point, for this particular neural network, is shown in Table \ref{table-adult}. \rchange{We can see that the race of this individual is influential in the decision of model, as are other features such as ``working hours" and ``work class". These two latter features seem to have an obvious causal relationship with the income, but influence of race should be questioned.}

\begin{table}[th]
\setlength{\tabcolsep}{5pt}
\caption{Difference in features for Adult dataset training point $\#53$ and its closest flip point.}
\label{table-adult}
\vskip 0.15in
\begin{center}
\begin{small}
\begin{sc}
\begin{tabular}{  p{2.1cm}  p{2.0cm}  p{3cm}  }
\toprule
Data & Input $\#53$ in training set & Closest flip point  \\
\midrule
Capital-gain    & 0 & 625  \\
\midrule
Capital-loss  & 1,902 & 1,862  \\
\midrule
Hours-per-week  & 60 & 59.8  \\
\midrule
Race  & White & Asian-Pac-Islander  \\
\midrule
Work class  & Private & State-gov  \\
\midrule
Marital Status  & Married-civ-spouse & Married-AF-spouse  \\
\bottomrule
\end{tabular}
\end{sc}
\end{small}
\end{center}
\vskip -0.1in
\end{table}

We can also constrain selected features when computing flip points. For example, we can ask for the closest flip point corresponding to a person with the same gender or race, or with a different gender or race. This enables us to investigate gender/racial bias in the output of the neural network.

\rchange{
{\bf Flip points reveal patterns in how the trained model treats the data.}
As an example, we consider the effect of gender (Male, Female) in connection with the family relationship (Wife, Own-child, Husband, Not-in-family, Other-relative, Unmarried) for individuals that have income ``$<=50$K". For this model, 89\% of data points in that income category have the same gender as their closest flip points, while 11\% have switched from Female to Male, and 0.2\% have switched from Male to Female. This shows that being Male is moderately helpful in being labeled ``$>50$K" by the model. But, as we will see later, education is the most influential feature for flipping to the high income category.

For the same income category, we also observe that for 2.5\% of the flip points, the family role switches from Husband to Wife, while a third of those have simultaneously switched from Female to Male. This reveals that the trained model considers both the family role of Wife and the gender of Male helpful for having high income. The switch from family role of Wife to Husband is absolutely rare among the flip points. 
}

{\bf PCA on the flip point directions identifies influential features.} 
Another important interpretation question is which features in the input  are most influential in the decisions of the network. To answer this kind of question, we use PCA on the matrix of directions between inputs and their flip points. Here, we discuss some of the insights that are obtained.

Consider the subset of directions that flip a ``$<=50$K" income to ``$>50$K". The first principal component reveals that, for this neural network, the most prominent features with positive impact are having a master's degree, having capital-gains, and working in the private sector, while the features with most negative impact are having highest education of Preschool, working without-pay, and having capital-loss.  Looking more deeply at the data, RR-QR decomposition of the matrix of directions reveals that some features, such as having a Prof-school degree, have no impact on this flip.

PCA on the directions between the mistakes in the training set and their closest flip points shows that native country of United States has the largest coefficient in the first principal component, followed by being a wife and having capital-gain. The most significant features with negative coefficient are being a husband and native countries of Cambodia and Ireland. These features can be considered the most influential in confusing and de-confusing the neural network.

PCA on the direction vectors explains how our neural network is influenced by various features. It thus enables us to calculate inputs that are mistakenly classified, for adversarial purposes. 

\rchange{
{\bf Flip points can deal with flaws and can reshape the model.}
Here as an example, we try to change the behavior of the trained model towards the individuals with country of origin ``Mexico". We observed that among various countries of origin, data points labeled ``Mexico" had the highest likelihood of a different country in their flip points. We consider all the data points with that country of origin that have a flip point with a different country. 82\% of those points have income ``$<=50$K". We generate closest flip points for all those inputs while constraining the country of origin to remain ``Mexico". We then add each generated flip point to the training set, using  the same label as the data point, and train a new model using the appended set. After performing PCA analysis on the directions to the new flip points, we observe that Mexico does not appear in any of the first 10 principal components, whereas it had a large value in the first principal component obtained for the original model. The accuracy of the trained model has remained almost the same (slightly increased by 0.05\%), confirming that we have achieved our goal. Using this kind of analysis, we can reshape the behavior of the model as needed.
}





\subsection{Wisconsin Breast Cancer Dataset} \label{sec-cancer}
Neural networks have shown promising results in identifying cancer \cite{agrawal2015neural}. As a simple example, we use the Wisconsin breast cancer database from the UCI repository which has 30 features extracted from digitized images of fine needle aspirate of  569 breast masses. We divide standard error features by their corresponding mean feature, and then normalize the mean and worst features between 0 and 1. The label is binary: ``malignant" or ``benign".

We randomly divide the dataset into a training set and testing set, consisting of 80\% and 20\% of data respectively. We achieve 100\% and 94.7\% accuracy on the training and testing sets, respectively. The average distance to the closest flip point is $0.022$ for the mistakes in the testing set and $0.103$ for the correct classifications in the testing set. The average distance is $0.106$ for correct classifications in the training set, very similar to the average distance in testing set. All of the mistakes have $\softmax$ score of at least 97.4\%. In fact, the average $\softmax$ for all the correct and wrong classifications are both more than 99\%. Again, the distance to the closest flip point is a reliable measure to identify classifications that are possibly wrong, while $\softmax$ score is not.

{\bf Flip points can be used to improve the model.}
What features in the input are most important? As an example, consider the first data point which is classified correctly as ``malignant" by the trained neural network. Its closest flip point differs mostly in features ``standard error of texture" and ``standard error of fractal dimension". 

We perform PCA on  the matrix of directions between each ``benign" input and its closest flip point, and look at the first principal component. The most prominent features that can flip the decision of the network to ``malignant" are ``standard error of radius'' and ``standard error of texture". Similarly, the most prominent features to flip a ``malignant" decision to ``benign" are ``standard error of texture" and ``worst area". 

 A clinician can use this information to validate the trained neural network as a computational tool. The information also enables the designer of the neural network to work with a clinician to rescale the data to emphasize features believed to be over- or under-emphasized by the current model and to provide better classifications.

\section{Conclusion}
We studied the problem of neural network interpretation and proposed methods that can interpret a trained neural network.

\begin{enumerate}
    \item We used flip points to investigate the boundaries between the output classes of any trained neural network. For any input to a neural network, we defined and solved optimization problems in order to find the closest flip point to each of the output boundaries. This provided accurate explanations about changes in the input that can flip the output from one class to another.  
    
    \item The distance of an input to the closest flip point proved to be a very effective measure of the confidence we should have in the correctness of the output, 
    much more reliable than $\softmax$ score. Moreover, this distance enables us to identify most and least influential points in the training data.

    \item PCA analysis identified the most influential features in the inputs. Also, for each output class, PCA identified the directions and magnitudes of change in each of the features that can change the output.
    
    \item  By computing relevant flip points, we created synthetic data and used it to boost the accuracy of a neural network. We also demonstrated how the synthetic data can be designed to adversarially affect a trained neural network by altering the output boundaries of the network. Our approach can be effective in identifying corrupt data, and also in generating corrupt data to attack a deep learning model.
    
    \item The distance and direction to the nearest flip point, coupled with a practitioner's knowledge of the measurement uncertainty in each of the features, can provide insight into whether the classification is unique or ambiguous.

\end{enumerate}

The computation and use of flip points greatly improves the interpretability of neural networks and enhances their use in applications where they have proven to be 
highly useful.

However, we note that flip points exist for any model, not just neural networks, and can be defined in a model-agnostic way as the closest point to a particular input that changes the model's decision or classification. We expect that flip points can be useful in providing insight and interpretation of a variety of types of models.





\bibliography{flip}
\bibliographystyle{icml2019}

\clearpage
\setcounter{table}{0}
\renewcommand{\thetable}{B\arabic{table}}
\renewcommand{\thealgorithm}{A\arabic{algorithm}}

\dchange{
\section*{Appendix A: How flip points are computed}

\rchange{{\bf General approach. }}
If the activation function is differentiable (e.g., $\erf$), we can make use of its gradient in solving the optimization problems we have introduced. Otherwise, subgradients can be used, but this can make the optimization algorithms more costly. 

The gradient of the outputs of the network with respect to its inputs is a {Jacobian matrix} when the network has more than one input feature and more than one output class. We compute this gradient analytically, which is generally more efficient and reliable than using finite differences.
Analytic computation of the gradients is possible for many different kinds of network architectures, including feed-forward, convolutional, and residual networks, assuming that the network does not contain non-differentiable elements such as non-differentiable activation functions or max pooling. 

For the feed-forward networks used in this work, the computation of the gradient is analogous to the back-propagation approach commonly used to compute the gradients with respect to the training parameters of the networks \cite{rumelhart1988learning}.

Using the gradients, we minimize  \eqref{eq-2o-obj} subject to the constraints mentioned \echange{in Section~\ref{sec_define_flip}} in order to find the closest flip point. The problem is non-convex, so standard optimization software   generally computes a local minimizer but not necessarily a global minimizer. In the case of inputs with discrete features, we can add the discrete constraints to the problem or add regularization terms to the objective function using the techniques described by \citet{nocedal2006numerical}.

\echange{Our} optimization problem can be considered a generally solvable problem using \echange{off-the-shelf} methods available in the \echange{literature}.
\echange{However, difficulties sometimes arise in solving nonlinear non-convex optimization problems, and therefore it is beneficial to design an optimization method tailored to our particular problem. To illustrate this, we first specify our network and then our optimization algorithm.}
}

In our notation, vectors and scalars are in lower case and matrices are in upper case. Bold characters are used for vectors and matrices, and the relevant layer in the network is shown as a superscript in parenthesis. Subscripts denote the index for a particular element of a matrix or vector. Iterations of the algorithm are denoted by superscripts. 

\begin{figure}[h]
\includegraphics[width=1\columnwidth]{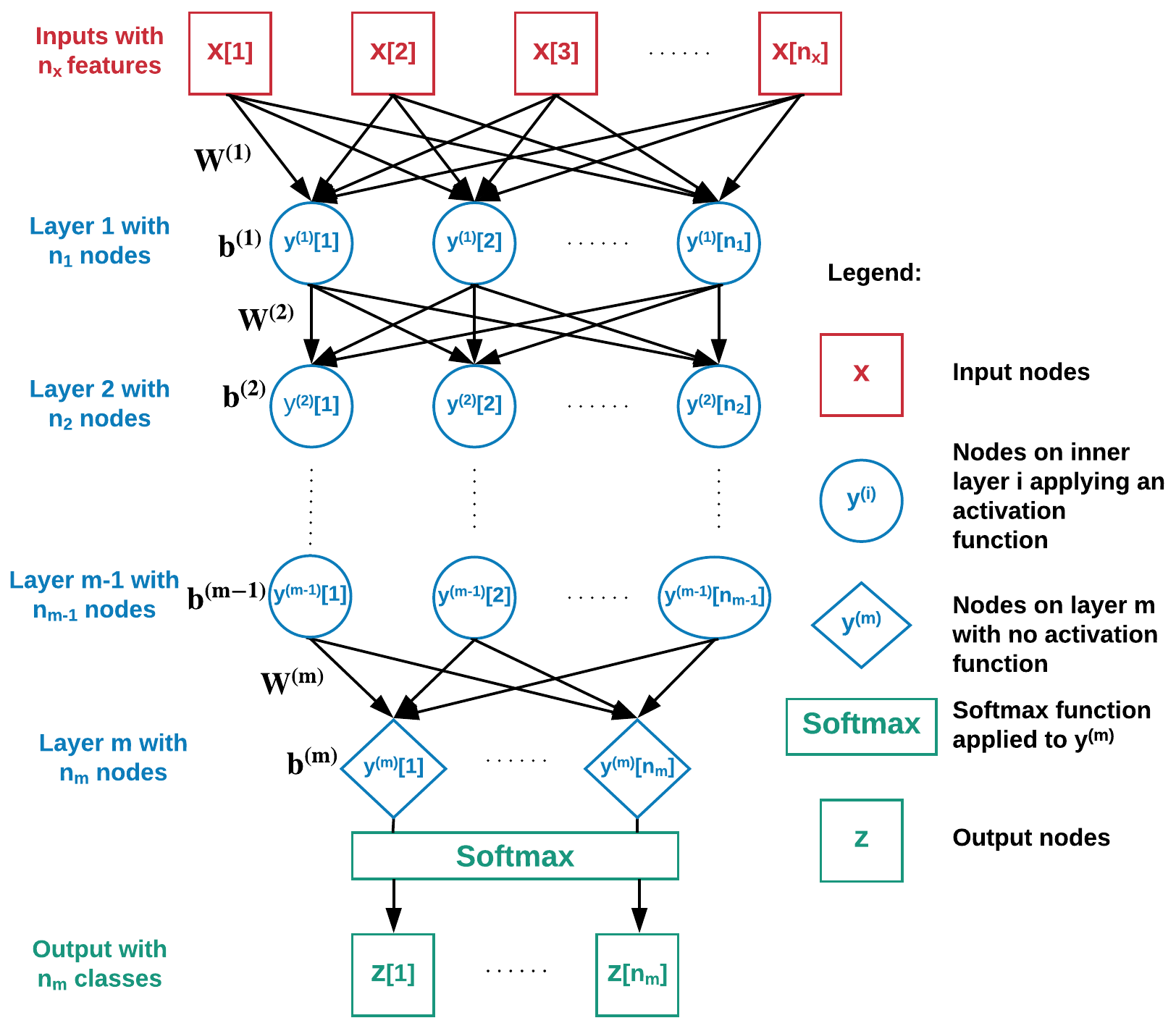}
\centering
\caption{Sketch of a prototype feed-forward neural network $\mathcal{N} $ with $n_x$ inputs, $m$ layers, and $n_m$ outputs.} 
\label{fig_network}
\end{figure}
\echange{\bf Our neural network.}
\echange{We specify the neural network $\mathcal{N}$  shown in Figure~\ref{fig_network} by weight matrices $\bs{W}^{(k)}$ and bias vectors $\bs{b}^{(k)}$ for each layer $k=1,\dots,m$. The output of layer $k$ in the network is denoted by $\bs{y}^{(k)}$.}

\begin{figure}[h]
\includegraphics[width=1\columnwidth]{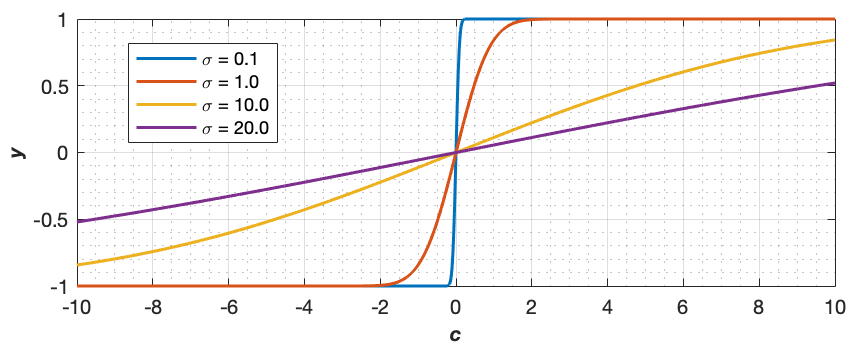}
\centering
\fchange{\caption{Shape of $\erf$ function as $\sigma$ varies.} }
\label{fig_activ}
\end{figure}

The activation function used in the nodes is the error function
\begin{center}
$y = activation(c|\sigma) = \erf(\frac{c}{\sigma}) = \frac{1}{\sqrt{\pi}} {\bigintss}_{\hspace{-3pt}-\frac{c}{\sigma}}^{+\frac{c}{\sigma}} {e}^{-t^{2}} { d}t,$
\end{center}
where $c$ is the result of applying the weights and bias to the node's inputs. The tuning parameter $\sigma$ is constant among the nodes on each layer and is optimized during the training process. Hence, for the whole network, we have a vector of tuning parameters, $\bs{\sigma}$, where each element of it corresponds to one hidden layer in the network. 
While $\erf$ is not a very common choice for activation function, it has been shown that its performance in terms of accuracy is comparable to other activation functions \cite{ramachandran2017searching}. \echange{We note that when $\sigma$ is small, then the activation function resembles \fchange{a step function}, while when $\sigma$ is large, it resembles \fchange{a linear function}, as shown in Figure \ref{fig_activ}, so $\erf$ captures the behavior of popular activation functions while preserving differentiability.}

Using the $\erf$ activation function, the output of the first inner layer for input $\bfx$ is
 $$\bs{y}^{(1)} = \erf \Big(\frac{\bfx \, \bs{W}^{(1)} + \bs{b}^{(1)} }{\sigma_1}\Big).$$
\echange{For the hidden layers,}
$$\bs{y}^{(i)} = \erf\Big(\frac{\bs{y}^{(i-1)} \, \bs{W}^{(i)} + \bs{b}^{(i)} }{\sigma_i} \Big).$$
\echange{Finally, }
$$\bs{y}^{(m)} = \bs{y}^{(m-1)} \bs{W}^{(m)} + \bs{b}^{(m)},$$ and the output of the network is
$$\bs{z}(\bfx) = \softmax \big( \bs{y}^{(m)} \big).$$

\rchange{\echange{\bf Optimization framework. }
We recommend a homotopy method for calculating the closest flip points. Here, we briefly explain its framework in the context of \echange{our network}. Our method can be easily generalized to neural networks with different architectures, such as convolutional and residual networks. \echange{The homotopy algorithm applies an optimization module to a series of networks.}
}

{\bf Optimization module.}
We \echange{define the numerical process of computing the closest flip point $\flip$ to an input $\bfx$ between classes $i$ and $j$ by the function $\mathcal{F}$:}
\begin{center} 
$\flip\pij = \mathcal{F}(\bfx,\mathcal{N},\bf{x}_0,\mathcal{C},$ $i,j),$
\end{center}
\echange{The inputs include the trained neural network $\mathcal{N}$, the starting point $\bf{x}_0$, and the constraints  $\mathcal{C}$. }
As a general practice and based on our numerical experiments, an interior-point algorithm can be considered a good choice, as it is known to be successful in solving constrained, nonlinear, non-convex optimization problems with high dimensional variables \cite{nocedal2006numerical}. \echange{This can be used in conjunction with a branch-and-bound algorithm for discrete variables.}

\echange{Ideally, $\mathcal{F}$ efficiently finds the closest flip point for our network, possibly using the input $\bfx$ as the starting point. If this fails, then we use a {\em homotopy method}, starting by applying $\mathcal{F}$ to an easier network and gradually transforming it to the desired network, each time using the previously determined flip point as our starting point for $\mathcal{F}$. We now discuss the family of networks used in the homotopy.}

{\bf Homotopy algorithms.} 
Our homotopy method, defined by Algorithm \ref{alg_homotopy}, \echange{begins with a neural network for which \fchange{$\bfx$ is a flip point}
, and then computes flip points for a series of networks, gradually transforming to the original network, using the closest flip point found at each iteration as the starting point for the next iteration.} This way, the algorithm follows a path of \echange{flip points} \fchange{starting from $\bfx$,} until it finds the closest flip point \fchange{to $\bfx$} for the original network.

\gchange{The initial neural network used in the algorithm is the same as the original network except that it has parameters $\bs{\sigma}^h$ for the $\erf$ and  $\bs{b}^{h(m)}$ for the bias on the last layer. These are computed in Algorithm \ref{alg_sigma}, discussed below.}

\echange{The parameter} $\eta$ defines the number of iterations that Algorithm \ref{alg_homotopy} uses to transform the network back to its original form. A large $\eta$ \echange{means that each neural network is a small change from the previous one, so the starting point is close to the solution. A small $\eta$ means that only a few optimization problems are solved, but each starting point may be far from the solution. We want to perform enough iterations so that the global minimizer is found, but we also want to keep the computational cost low.}
\echange{We} have achieved best results \echange{with}  $\eta$ ranging between 1 and 10. \gchange{Choosing $\eta = 1$ is equivalent to not using the homotopy algorithm and directly applying  $\mathcal{F}$ to the original network with starting point $\bfx$.} 

\begin{algorithm}[t]
\caption{Homotopy algorithm for calculating closest flip \echange{point}}
\label{alg_homotopy}
\textbf{Inputs}: $\mathcal{N}$, $\bfx$, $\eta$, $\tau$, $\mathcal{C}$, $i$, $j$ \\
\textbf{Output}: Closest flip point to $\bfx$
\begin{algorithmic}[1] 
\STATE Compute $\bs{\sigma}^h$ and $\bs{b}^{h(m)}$ using Algorithm \ref{alg_sigma} with \echange{inputs} ($\mathcal{N}$, $\bfx$, $\tau$, $i$, $j$)
\STATE $\flip{}^{,0} = \bfx$
\FOR{$k = 1$ to $\eta$}
	\STATE $\bs{\sigma}^k = \bs{\sigma}^h + k (\frac{\bs{\sigma}^{\mathcal{N}} - \bs{\sigma}^h}{\eta} )$
	\STATE $\bs{b}^{k(m)} = \bs{b}^{h(m)} + k (\frac{\bs{b}^{\mathcal{N}(m)} - \bs{b}^{h(m)}}{\eta} )$
	\STATE Replace $\bs{\sigma}^k$ and $\bs{b}^{k(m)}$ in $\mathcal{N}$, to obtain $\mathcal{N}^k$
	\STATE $\flip{}^{,k} = \mathcal{F}(\bfx,\mathcal{N}^k,\flip{}^{,k-1},\mathcal{C},i,j)$
\ENDFOR
\STATE \textbf{return} $\flip{}^{,\eta}$ as the closest flip point to $\bfx$
\end{algorithmic}
\end{algorithm}

\begin{algorithm}[t]
\caption{Algorithm to \fchange{transform the network for the Homotopy algorithm} }
\label{alg_sigma}
\textbf{Inputs}: $\mathcal{N}$, $\bfx$, $\tau$, $i$, $j$  \\
\textbf{Output}: $\bs{\sigma}^h$ and $\bs{b}^{h(m)}$
\begin{algorithmic}[1] 
\STATE $\gamma = \sqrt{log(\frac{2}{ \tau \sqrt{\pi} })}$
\STATE $\bs{y}^{(0)} = \bfx$
\FOR{$k = 1$ to $m-1$}
	\STATE $\sigma_k^h = \max( \frac{2}{\sqrt{\pi}} , \frac{1}{\gamma} \, \| \bs{y}^{(k-1)} \bs{W}^{(k)} + \bs{b}^{(k)} \|_{\infty} )$
	\IF{$\sigma_k^h > \frac{2}{\tau\sqrt{\pi}}$}
		\STATE $\bs{c} = \bs{y}^{(k-1)} \bs{W}^{(k)} + \bs{b}^{(k)}$
		\FOR{$t = 1$ to $n_k$} 
			\STATE $\sigma_{k,t}^h = \max( \frac{2}{\sqrt{\pi}} , \frac{1}{\gamma} \, c_t )$
		\ENDFOR 
	\ENDIF
	\STATE $\bs{y}^{(k)} = \erf(\frac{\bs{y}^{(k-1)} \bs{W}^{(k)} + \bs{b}^{(k)}}{\sigma^h_k} )$
\ENDFOR
\STATE $\min\limits_{\bs{b}^{h(m)} } \| \bs{b}^{h(m)} - \bs{b}^{\mathcal{N}(m)} \|_2$\, , subject to: \\ \label{alg2_b}
	{\small (1) }$\bs{y}^{(m)} = \bs{y}^{(m-1)} \bs{W^{(m)}} + \bs{b}^{h(m)}$, \\
	{\small (2) }${y}^{(m)}_i = {y}^{(m)}_j$, \\
	{\small (3) }$\forall~l~\ne~i,j \; | \; {y}^{(m)}_i > {y}^{(m)}_l$
\STATE \textbf{return} $\bs{\sigma}^h, \bs{b}^{h(m)}$
\end{algorithmic}
\end{algorithm}

The initial transformation of the network is \echange{performed} by Algorithm \ref{alg_sigma}, pursuing two goals\fchange{, first, bounding the flow of gradients through the layers of the network by changing the value of tuning parameters (lines 1 through 12), and second, changing the bias parameters in the last layer of the network \gchange{so} that $\bfx$ is a flip point for the transformed network (line 13).}

\gchange{The tuning parameters for the original network are} $\bs{\sigma}^{\mathcal{N}}$, and $\bs{\sigma}^h$ denotes the transformed parameters computed by Algorithm \ref{alg_sigma}. Similarly, $\bs{b}^{\mathcal{N}(m)}$ and $\bs{b}^{h(m)}$ denote the original and transformed bias in the last layer of the network.

\fchange{
By changing $\bs{\sigma}^{\mathcal{N}}$ to $\bs{\sigma}^h$, we \gchange{try to control the magnitudes of the} gradients of output with respect to inputs. The hierarchy of neural networks can cause the gradients to vanish and/or explode through its layers, which could lead to a badly scaled gradient matrix and eventually an ill-conditioned optimization problem, \gchange{and we would like to avoid this. } For flip point computation, we are concerned about the gradients of outputs with respect to inputs, while in neural network literature, this issue of ``vanishing and exploding gradients" usually concerns the training process and the gradient of the loss function with respect to the training parameters \cite{bengio1994learning,hanin2018neural}. In both cases, the ``vanishing and exploding gradients" phenomenon can be studied by investigating individual matrices in the chain rule \gchange{formulation} of the gradient matrix. 

To compute the $\bs{\sigma}^h$, we trace the $\bfx$ as it flows through the layers of the network. As the input reaches each hidden layer, before applying the activation function, we tune the corresponding element of $\bs{\sigma}^h$, \gchange{so} that the absolute values of the gradients of the output of each neuron, with respect to neuron's input, is greater than or equal to $\tau$, and less than or equal to 1. 
In our numerical experiments, we have used different values of $\tau$ ranging between $10^{-5}$ and $10^{-9}$.
}

\fchange{
In Algorithm \ref{alg_sigma}, line 1 computes a scalar $\gamma$ such that \gchange{the derivative of the} $\erf$ is equal to $\tau$. Lines 3 through 12, tune the $\sigma$, layer by layer, starting from the first layer and ending at the last hidden layer. Line 4 bounds the \gchange{individual gradient} between $\tau$ and 1. Choosing the $\sigma_k^h > \frac{2}{\sqrt{\pi}}$ ensures the gradients of neurons are upper bounded by 1. This relationship can be easily derived by setting the maximum derivative of $\erf$ equal to $\tau$.

Choosing $\sigma_k^h \geq \frac{1}{\gamma} \, \| \bs{y}^{(k-1)} \bs{W}^{(k)} + \bs{b}^{(k)} \|_{\infty} $ can potentially  make the gradients of all the neurons in layer $k$ lower bounded by $\tau$. Sometimes, this might not be possible to achieve for all the neurons in a layer, if we obtain $\sigma_k^h > \frac{2}{\tau\sqrt{\pi}}$. In such situations, we calculate the $\sigma_k^h$ separately for each neuron on that layer (lines 5 through 10), and use a non-uniform $\sigma_k^h$ in the homotopy algorithm. Line 11, computes the output of each layer after the $\sigma$ is tuned for that layer.

Since our activation function is $\erf$, we can effectively control the gradients and make them bounded. The maximum gradient of $\erf$ is at point zero, and by moving away from zero, its gradient decreases monotonically, until it asymptotically reaches zero. This boundedness and the monotonicity of both the $\erf$ and its gradient are helpful features that we leverage in our homotopy method. When using activation functions other than $\erf$, we have to avoid exploding and vanishing gradients, depending on the properties of the activation function in use. 

}
\fchange{
By changing $\bs{b}^{\mathcal{N}(m)}$ to $\bs{b}^{h(m)}$, computed at line 13 of Algorithm \ref{alg_sigma}, the input $\bfx$ actually becomes a flip point for the transformed network. Having a starting point that is feasible with respect to flip point constraints considerably facilitates the optimization process. The optimization problem on line 13 of the algorithm is \gchange{a convex quadratic programming problem and can be solved by standard algorithms.}}

\newpage

\section*{Appendix B: Information about neural networks used in our numerical examples}

Here, we provide more information about the models we have trained and used in Section \ref{sec_results}. 

We have used fully connected feed-forward neural networks with 12 hidden layers. The number of nodes for the models used for each data set is shown in Table \ref{table_nodes}. \rchange{The activation function we have used in the nodes is the error function, as defined in Appendix A. We have also used $\softmax$ on the output layer, and cross entropy for the loss function.}

\begin{table}[h]
\setlength{\tabcolsep}{5pt}
\caption{Number of nodes in neural network used for each data set.}
\label{table_nodes}
\vskip 0.15in
\begin{center}
\begin{small}
\begin{sc}
\begin{tabular}{  p{2.cm}  p{1cm}  p{1cm} p{1cm}  p{1cm} }
\toprule
Data set & MNIST & CIFAR-10 & Adult & Cancer (WBCD)  \\
\midrule
Input layer & 100 & 2304 & 107 & 30  \\
\midrule
Layer 1 & 500 & 400 & 100 & 40  \\
\midrule
Layer 2 & 500 & 400 & 100 & 20  \\
\midrule
Layer 3 & 500 & 400 & 100 & 15  \\
\midrule
Layer 4 & 400 & 350 & 80 & 10  \\
\midrule
Layer 5 & 300 & 300 & 60 & 5  \\
\midrule
Layer 6 & 250 & 250 & 50 & 5  \\
\midrule
Layer 7 & 250 & 250 & 50 & 5  \\
\midrule
Layer 8 & 250 & 250 & 50 & 5  \\
\midrule
Layer 9 & 200 & 200 & 40 & 5  \\
\midrule
Layer 10 & 150 & 150 & 30 & 5  \\
\midrule
Layer 11 & 150 & 150 & 30 & 5  \\
\midrule
Layer 12 & 100 & 100 & 20 & 5  \\
\midrule
Output layer & 10 & 2 & 2 & 2  \\
\bottomrule
\end{tabular}
\end{sc}
\end{small}
\end{center}
\vskip -0.1in
\end{table}

\end{document}